# AI for the Generation and Testing of Ideas

## Towards an AI Supported Knowledge Development Environment

Ted Selker, Ph.D.

Selker Design Research
Palo Alto CA USA
Ted.Selker@gmail.com

## ABSTRACT

New systems use Machine Learning (ML) to cull through large sources of knowledge and create versatile Large Language Models (LLM). LLM Transformer Models recognize context and predict sequential information in writing and other communication forms. Generative Artificial Intelligence (AI) uses Transformers to create textual or visual output that can mimic human responses to prompt requests. Generative AI posits one or several contextually-plausible solutions for a user to consider. Traceability of ideas is not yet supported by generative AI. Traceable provenance to show where things come from is a powerful service that search engines provide.

The generative-AI-story-presentation style has been well-received. People learn from stories. But early ChatGPT efforts struggled with truth, reference, calculation, and things like an actual map of a place. Referenceable locations and links to apps currently seem to be better served through the link-focused search paradigms we have been using for 20 years. Deploying deeply believable solutions is more than simulating their contextual relevance, as generative AI does. Hybrid scenarios can bring together the imagination of generative AI and the provenance of internet sources to improve people's use of the Internet's knowledge.

Taken as drafts, generative AI's contextually-proposed material gets people thinking. We consider alternative ideas to form the final version or action in the world. This paper considers scenarios for requesting information. The paper then describes how generative AI can increase idea creation by removing human ego from the discussion. We describe how search can deepen our testing of facts, logic, and context. The person completes the evaluation of generated ideas, choosing and using them in real solutions. This paper proposes a plausible knowledge-worker system, Generate And Search Test (GAST), that lets individual seekers efficiently create solutions that here-to-fore required the best collaborations of the best experts.

This paper is part of considering ways that existing and emerging knowledge paradigms can help us. AI can interactively help us consider new knowledge approaches to achieve valuable goals.

## CCS CONCEPTS

• **Computing Methodologies →Artificial Intelligence →Natural Language Processing • Computing Methodologies →Document and Text Processing → Document Preparation • Information Systems→ Information Interfaces and Presentation → User Interfaces.**

## KEYWORDS

AI; Generative AI; Large Language Model, Machine Learning; Transformer Model, User Experience; Search

## 1 Knowledge

People spend their lives collecting knowledge and living by their understanding of topics and areas they care about. They consider the building blocks of existing solutions and sometimes imagine new ones they might like to try. When lucky, people are successful at creating solutions and are allowed to share them to help others.

Artificial Intelligence is a soft term that carries the baggage of simulation of human intelligence and a history of focusing on computer reasoning, representation, and learning [19]. The Machine Learning (ML) part has become a powerhouse of useful classification work being used in many industries. Databases and early search engines used humans to hand-classify knowledge [1,5]. Today search system classification is done with the most sophisticated ML technology in the world to respond to the most complex requests [2].

Search connects keywords with Boolean operators and presents lists of solutions relative to knowledge, and also opportunities relative to these queries [11]. They draw from an AI representation of all knowledge on the internet topresent lists of responses, information, and opportunities relative to these queries. Some search technology uses AI to create web pages and knowledge-source alternatives to the official ones. Searching for a restaurant, for example, often brings up a box filled with information about the restaurant that is put together by the search engine, not the restaurant itself, offered by the restaurant or other sources. There is huge value in knowing that knowledge is reliable. Search systems find knowledge in their primary sources. Search engines then have value for discovering, checking, and expanding one's conceptions, while also basing this work on traceable provenance.. Alternative ways of searching and browsing for information have evolved as well.

A popular direction in AI, is classifying large knowledge corpora of language or interpreted visual examples into Transformers, creating so-called LLMs [8,3]. Generative AI starts from human questions. It uses Transformers to respond to these questions. The solutions are an amalgam of knowledge it has compiled from the corpora it has assimilated, knowledge of how to put together a solution (story, poem, image, computer program, etc.), and knowledge of how to imagine and promote ideas in such forms.

Generative AI demonstrated with OpenAI's ChatGPT, for example, lets people present questions and clarifications in convenient Natural Language form [18]. The user experience of asking questions and getting plausible but test-worthy responses is familiar to human experience. Based on the goals a request presents, ChatGPT's responses focus on creating a structurally-and rhetorically-plausible presentation, not on delivering vetted factual responses [18].

Luckily, people are used to not having all stories correct or complete. The value of discovering, evaluating, and expanding one's conceptions is best checked with traceable provenance. Source provenance has to be made apparent to continue to keep the facts, not rhetoric, in distinguishing reliable sources from fabricated stories. Search is valued for its ability to produce factual responses.

## 1.1 Knowledge is Social

Peoples' reputations can rely on coming up with ideas and getting credit for them. The fight for credit can be ruthless and fierce. Instead of having and evaluating many ideas, someone might decide to get social accolades by getting everyone on their side early. Sometimes there is a person in a group that is much more capable, knowledgeable, and decisive than others. This is fine if the meeting is to disseminate the special result from the special person as an oracle. This tendency to be the oracle might not be the best way of solving any problem. The computer can also act as an oracle but we don't have to cede our position to get it to do something for us. We typically prepare to solve problems by learning for a few dozen years, then fine-tuning our knowledge when we have a new and similar problem to what we have considered.

But excellent social structures can exist where people feel proud of their collaborations with each other too. Presumably everyone in a decision has special knowledge and perspective that they bring to help find solutions. Best results can come from a group of collaborators thinking through ideas and considering alternatives to ones that are proposed. For the most objective results the elaboration of the ideas should be egoless. Sometimes it can happen with humans

Considering alternative ideas is an active process that starts with a deep knowledge about the problem, solutions, and appropriateness of the solution. For example, I was designing laptop computers at IBM, when the American Scientific expedition to Everest wanted to record biometrics for climbers. I was asked to support the expedition with laptops that would work at altitude. My research lab happened to be in the same building with people working on disk drive innovation. I got together with disk drive design experts to discuss a new disk drive that would work. Their recent relevant work had been around helping disks work in airplanes. Specialized knowledge was critical in this case, but for most problems the internet can now find even specialized knowledge

In spite of their familiarity with using disks in airplanes, using them at Everest brought out many new ideas. The conversation was, as the best brainstorms are, people contributing and evaluating many different alternative ideas. In each case, we used the group's shared experience and knowledge to with a shared understanding of its tradeoffs. For example, we considered inventing special disk heads with adjustable ailerons that hat would auto adjust for elevation. Like the conversation, searches allow us to find at diverse possibilities; generative AI naturally presents diverse possibilities and takes a crack at prioritizing them.

After considering the results of an exploration we can find ourselves reprioritizing results. In this case, I added the constraint that the computers had to be ready in 3 months. At that point, there was a collective groan, as we looked for simple solutions. Out-of-spec disks were looked for on the manufacturing line that wouldn't work at sea level. Within a week 5 disks were found that would only work at high elevation. New constraints can be expressed as a longer prompt in Generative AI or a set of search tabs and links in search.

This was a bunch of people open to many solutions and they were delighted to collaborate. This example of collaboration is the sometimes-rare combination of many sources of ideas and a willingness to cast aside all but the best. Many generative AI experiences start with wrong answers that need more constraints too. The many sources of knowledge needed to solve it and the interplay between people underscore how critical the back and forth with a generative AI solution might need to be to simulate such an experience.

Often people in an idea generation session aren't as open to new ideas as this story describes. We get stuck on one idea, knowing it isn't good enough. Calling it procrastination or writer's block or whatever, sometimes we can't get started. Requests are easier to come by than solutions. Today out-of-box ideas often come from the egoless computer as well as a trusted group. Trust is hard, so collaboration is delicate, as it requires people to shelve their need to win accolades.

Generative AI gives a narrative solution that needs to be considered and checked for viability. This gives the user something to do. A sequence of searches may have brought ideas, but the results come mixed in with many distractions of 'out of context' content and require several searches to refine. If for no other reason than not coming from another person that you might be competing with for attention, people tend to delight in, instead of resisting, generative AI suggestions.

As alluded to above, one must separate ownership from quality. We become frightened of sharing our problem or solution if it will contribute to someone else dominating the meeting. One must be comfortable with other people owning ideas. One must trust and respect the competence of others in the group and also be patient with alternative ideas, which may slow the deepening of one idea. Generative AI's most potent power might be to generate and promote an idea that people will consider without authority or ownership. The social element is likely the most difficult part of an idea - a problem that doesn't exist with machine-generated suggestions.

The process of constructing an understanding might be from experiences and self-learning. It might be from courses in school, a result of mentorship, or it might be from looking for knowledge on one's own. A few decades ago, people primarily looked for information in libraries and catalogs that came in the mail. More recently, we use search engines on specialized corpora or the internet to tap into comprehensive knowledge. For many kinds of problems, we now find that letting generative AI create a solution can be a shortcut that might be even more elucidating, even if it has inaccuracies. These inaccuracies come in the form of made-up facts the AI community has been calling hallucinations. That the hallucinations are not completely made up makes me call them fantasies. Fantasies are ideas that are interesting because they don't exist but fill in gaps or are part of a fun story. Fiction captivates people with fantasy. It starts start with plausible physics, world views, and occurrences. Generative AI uses its vast knowledge to come up with probable compelling ideas. Accepting these fantasies as knowledge is dangerous. This paper makes a case for tools that help us to separate knowledge from fantasy in generative AI systems.

## 1.2 An Evolution of Knowledge Tools

Technology has been helping us use language forever [7]. Writing instruments are always under improvement, starting with things that were carved into a surface such as sand, rock, or wood. Others added to the surface with chalk, and carbon, or left a stained trail with a colored material such as beet juice, blood, etc. Printing presses and, later on, printers, allowed writers to make copies to distribute or edit. Displays allowed writers to view and change their writing at will.

Standardizing spelling and syntax made it easier for people to read and understand each other's writing. The dictionary standardized spelling. Published in 1604, Robert Cawdrey's is said to be the first dictionary [6]. Cut & paste and word completion was introduced with Emacs. Cut & paste in text editors sped accurate transcription and flexibility in reorganizing thoughts. I did an automated global replacement of hundreds of instances of "the fucking" with "the" in my Ph.D. thesis draft to make it G-rated before showing it to readers.

Word-completion pulldown menus became an especially sought-after feature to speed typing in Japanese word processors. Computer spell check became a valuable tool for people using word processors starting in the late 1970s. Do What I Mean (DWIM) added searching for syntactic context and consistency to find places that used the same or similar language in the Interlisp-D computer program in the 1980s [20,4]. Its ideas are now used in most computer programs in Integrated Development Environments.

In 1981, the IBM EPISTLE [12] system demonstrated the possibility of improving authors' syntax. This was tested by IBM with writers at Vassar College. The creator of it, George Heidorn, left IBM to implement it in Microsoft Word [15]. The legibility improvements that it created have continued to improve. This was a watershed improvement for evaluating what people wrote using AI. At the time it was dependent on hand-built knowledge-based reasoning. Now 30 years later, this is so important that it has created an industry of grammar and legibility checking. Today such tools are created by much bigger and more robust statistical models.

It isn't even new to have computers put their results in natural language. In the 1960s, systems like Eliza [21], for example, produced compelling utterances. Systems like HearsayII in the 1970s allowed people to play chess by talking to a computer [10]. The first decade of the third millennium brought ubiquitous question-answering voice interaction in Siri, Alexa, Google Assistant [16,17], and others. The question-answering abilities varied by the ability for the voice to be recognized and access an index of available knowledge query types. Open AI's ChatGPT and others, like Google's BARD, prioritize solution flow and continuity over correctness of response [18]. Their input and output approach is tuned to user expectations of conversation.

## 1.3 Active AI Responses

As introduced above, the evaluation of ideas using computers has taken many forms. For so long, creativity-enhancing software consisted of computer "outliners" and "mind maps". No more. We search the world's knowledge on Google, published articles on Google Scholar, edited explanations on Wikipedia, etc. We turn to The Internet Archive for the things people have put up in the past but may have rescinded or abandoned.

When introduced, Siri, a knowledge-based verbal question-answering system described above, seemed like a breakthrough [16]. It started as SRI's knowledge-based office-assistant research in the new millennia. Siri was a spinoff that focused on commercial opportunities to buy things. When it was acquired by Apple it became a general voice question-answering system. This was followed by the less brittle machine-learning-based versions from Google and others. The results competed with search and were also used for smart control of a household. The ability to guide ChatGPT with style, qualities, and quantity of responses presents a completely new kind of interaction paradigm. The requests and responses focus on how what should be presented how. As described above it can present as a person might and revels in giving alternatives that encourage broadening the solution space.

## 1.4 Knowledge Interaction Scenarios

When asked "What is a quark?" ChatGPT answers:
"A quark is a subatomic particle and a fundamental constituent of matter. Quarks are elementary particles, which means they are not made up of smaller particles. Quarks come in six different types, known as flavors: up, down, charm, strange, top, and bottom.

Quarks are always found in combination with other quarks or with another fundamental particle called a gluon. When quarks are combined together, they form particles known as hadrons, such as protons and neutrons. The way that quarks interact with each other is governed by the strong nuclear force, which is one of the four fundamental forces of nature."

While Alexa also forms almost the same grammatically flowing response, there are so many differences between these tools. While both have a model of discourses, ChatGPT has a model of persuasion, making up plausible results and filling in with plausible ideas. ChatGPT is designed as a private conversation. Alexa uses speech-based input and output, responses are heard by everyone within earshot taking social space from the room. But it doesn't have the generative AI notion of filling in gaps with likely-sounding responses, as people do to make conversation.

Still, both of these platforms have limited corpus knowledge bases compared to the web. Updating the indices of search systems is an up-to-the-second and ongoing comprehensive activity, while updating generative AI systems is not yet that encompassing. Alexa and ChatGPT both feel less truthful and comprehensive than internet searches for different reasons.

## 2 New Scenarios for Knowledge Interaction

Search has changed the way we find, use, and acquire things. The multiple goals of search have moved it away from finding a perfect answer to some sort of negotiated set of results that the search engine and the user interact with. Early purpose-built search engines typically seemed limited. The search engine of today serves many goals giving:

- marketing results
- sales results
- website results
- video results
- scholarly results
- how-to results
- AI website-produced composite results, for businesses
- a menu bar allowing you to use your search query for shopping, images, videos, news, maps, books flights, and finance.

We have many threads of goals in our heads. The many goals of the search system mean that one searches and then browses related things, to find the part related to their goal for searching in the first place. To satisfy these threads and also to monetize, search brings in commercial opportunities.

A more special-purposed search engine might help us focus. Maybe the search engine shouldn't be promoting a new outfit, toolkit, or vacation opportunity when searching for a story's veracity and provenance. I once asked Marissa Mayer why Yahoo was presenting a married man with opportunities for breast enhancement and dating service advertisements. She said, "We tried to take them out, but the advertising opportunity makes a lot of money for Yahoo.". Focusing these systems only on the users' goals will improve people's productivity. With a subscription-based model, so far, ChatGPT is avoiding cluttering solutions with predatory results.

Even in 1998, when I taught a class at MIT on the future of technology, my colleague professor Jacobson projected a screen with a stream of Google search results for everything said in the class. It was done in a yes-and instigating way that added to the conversation. Now we have much better tools. Let's extrapolate some scenarios and the value they can bring.

### 2.1 Scenarios and their possible value

*2.1.1 Generative AI Can Feel Natural to Use.* Shortly after ChatGPT's initial release, I was at a gathering of octogenarians talking about their use of it. I was shocked at their initiative and sophistication. One of them read a competent poem in iambic pentameter using the language of Shakespeare talking about something current this person was thinking about. Another was complaining that all the quilting patterns she was able to solicit from it she saw as childlike and simplistic. We are no longer surprised by generative AI's relevance.

*2.1.2 Generative AI Can Provide Entertainment.* I was in a rainforest resort in Sylhet Bangladesh; a group of people were laughing so hard they were almost falling out of their chairs. They had asked ChatGPT to write a poem in the style of the most famous Bangladesh poet. One of them was singing it as a song in Bengali!

*2.1.3 Generative AI Can Generate Sophisticated Suggestions.* In creating a training set for a project, my team used ChatGPT to author apologies. In each case, after a few sentences, it repeated its structure and even statements. Still, its suggestions helped people write training utterances almost 10 times faster than before.

An especially telling example came from a friend writing a piece for a prominent periodical. He asked ChatGPT to tell him about disruptions in 1923 after the Spanish flu pandemic to compare with expected disruptions after the Covid pandemic in 2023. He fact-checked his source, and, yes, found some of them spurious. He used some of the ideas; the answers also inspired him to generate more ideas. He said that it allowed him to write an article in a fraction of the time it would have otherwise taken.

*2.1.4 Generative AI is Problematic as a Search Replacement.* A friend asked ChatGPT about the best recording Shirley Temple made; it said she never recorded.

"It's worth noting that Shirley Temple, despite her fame as a child actress in the 1930s, retired from acting at the age of 22 and did not release any songs during her career. The songs in the list above are fictional titles and are not associated with any actual songs or albums by Shirley Temple."

The follow-up question was focused on giving it more information; ChatGPT then contradicted itself, noting her many real recordings.

*2.1.5 Generative AI Can Create.* Various online support services like Stack Overflow allow programmers to learn from others' examples. Generative AI is now assembling all known code to help people program. More than just reviewing code, it is being used to write code with tools like Copilot. People can name algorithms, the programming language to be written in, and the output requested to receive working programs. By criticizing the result, they can get Copilot to fix bugs, change the approach, or even port the program to another language in seconds. In this way, the debugging cycle includes going through and fixing many little bugs, one at a time. It gives useful sample code; customizing the examples, on the other hand, can be an extensive activity of expressing specific requests as questions. Still, programming complex things by asking questions can feel like trying to drive a car from the back seat.

Creating compelling interactions with people is also important. People are used to conversation, but search does not support conversation or story. When generative AI brings back problematic responses and buggy programs, it becomes a conversation. Conversation may not be the best way to find facts or for that matter to analyze and solve bugs in a computer program. The value of injecting fact-checking and provenance findings is central to moving from conjecture to knowledge. The future of ideas and knowledge can be prompted by made-up stories, but they are only valuable when their veracity is established.

Criticizing and asking questions is not like creating. The critical questions of an art historian are not the creative act of an artist. The generative AI scenario responds to a request with constraints with creative output. But people find ease in the scenario, as recognizing a solution is easier than recalling one, and especially easier than creating one.

The field is busily exploring how people best communicate while producing useful products. People are using generative AI to help write letters, essays, articles, songs, speeches, computer programs, and more.

We spend our lives communicating to learn and do. A challenge we have is to create a quiver of knowledge and communication tools that people value and are productive with. The search paradigm has been transformative compared to the older internet tools like FTP, Archie, Veronica, and WAIS[9]. It is giving us access to the world's knowledge.

Search straddles various knowledge needs.  While newer versions of it are now aggregating information about specific enterprises, like restaurants or other businesses, it typically focuses on showing many alternative links that might be followed. Finding things is different than accumulating a solution. The search paradigm isn't designed to accumulate knowledge, crucially analyze discourse, or formulate solutions. We are at a threshold where the ease of all the many ways we use knowledge is enhanced by the corpus of all digitized knowledge.

Today's search engine is expected to respond appropriately to the full variety of reasons a person might want informational help.  It is expected to differentiate well enough to allow a person's needs to be viewable in a few seconds of search, but is that realistic?  Queries are given to a system that doesn't know the context; the background, sophistication, and goals of the request have not been a first-class participant in the search activity.

The following is a partial list of possible ways that a person comes to the search activity.

### 2.1.6   Many Information Access Needs Are Not Yet Supported by Generative AI:

1.  **A fact checker** is what we call ourselves when we are finding out where the information came from and making sure it is correct.  We are not looking to buy things. Provenance is everything here; the result must be correct and traceable.
2.  **A scientist** is what we call ourselves when we are looking for knowledge to make and test hypotheses in an area. We are not just looking for confirmation of information but related knowledge from the known and vetted scientists in their area, but also in plausible up-and-coming groups and people, and also from outliers.  Knowledge of colleagues and bibliographies of scientists' papers are typical go-to sources.
3.  **A recreational intellectual** is what we call ourselves when we are interested in being current in an area where we probably won't contribute original work. We are looking for the obvious and known facts, be it trivia or deeper factual information. This is a simple search. Since intellectual excursions are often done as entertainment, search engines might appropriately suggest other entertainment support, like food and drink.
4.  **A Do It Yourself (DIY)** person is what we call ourselves when we are trying to fix or build something, without being a professional or without using professional tools. We might be looking for how-to information to solve our problem without having to contact an expert or take a class. This has monetary opportunities for education, materials, and services.
5.  **A hobbyist** is what we call ourselves when we are making a practice of some activity that might include others doing it too. We might be looking for people and ways to exercise and be known for our interests. Visual and video information is easily accessed with standard search for many of the needs here.
6.  **A constructor** is what we call ourselves when we are creating something new or impressive [14].  We are trying to find methods, materials, and analysis to support creation. This has been successful with standard search tools.
7.  **The professional** is what we call ourselves when we are using best-of-breed tools to do best-of-breed work. We are trying to assess best practices, best tools, job, career opportunities, and enrichment. This is a contextual search.
8.  **The merchant** is what we call ourselves when we are selling or buying things. We are looking to find markets-places and people that will want our goods and services. Search websites get to specialized information and interaction quickly here.
9.  **The consumer** is what we call ourselves when we are trying to acquire something. We are evaluating known and unknown needs and opportunities to dream about them and fulfill them. The standard search seems to still use collaborative filtering.  Specialized web sites are more quickly accessed with search here.

A variety of explorations performed to compare search with ChatGPT 4 for the above use cases informed the following five conclusions.

1.  Search gave better experiences for queries that could be explained with spatial relations or responses.
2.  Queries that had very specific responses (like how to fix a fender well in a 2010 Prius) were much more correct and helpful with search.
3.  Explorations for commonly available information yielded similar results with the two approaches
4.  ChatGPT results were easier to look through as it presented in narrative form and didn't have the clutter of advertising and other search goals not intended in the question.
5.  ChatGPT had the added advantage of showing its prioritization results explicitly.

### 2.1.7   Generative AI Presents Information in Ways People Are Used To：

1.  **A knowledgeable domain expert** is what we call ourselves when we have scoured any topic.  We are like scientists but not looking to start experiments. We want to get information that respects and reinforces our knowledge as well as extends it. Search works well for identifying specific information, finding famous people in an area, etc. A generative AI system should also be able to broaden an expert's understanding.
2.  **A student** is what we call ourselves when we are just trying to understand techniques and get a broad understanding of fields. We might be trying to demonstrate proficiency through the demonstration of facts, procedures, analysis, or work product. Generative AI has been demonstrated to be extremely valuable for being part of this kind of solution.
3.  **A communicator** is what we call ourselves when scripting presentations. We are scripting something to share, educate,

or document. Generative AI is changing the world with its new paradigm for these things.

Making a variety of searches for the above cases showed at least two big benefits for generative AI.

1. The ability to continue to follow up, expand and focus a question is demonstrated well in examples such as a student using it as a tutor.
2. Results coming as a presentation is demonstrated well for a communicator whose result will be a presentation.

Generalized search engines like Google have been trying to cover all the uses a person might have for it, reducing the need for specialized tools. Trying to cater to any information need, they produce lists and web pages. Still, special-purpose search engines or search engine settings can be useful for looking through refereed articles, different media, visualization, domain access, and specialized information.

The emerging ChatGPT paradigm lets a user add any amount of context for a query naturally as they define and redefine their prompt request. For many people, the discursive input of ChatGPT is turning out to be easier to master than the Boolean-esque languages search engines depended on for user input. Formulating a question with keywords is a search. Formulating a question in a narrative for generative AI is an act of exploration that might make us more open to diverse answers.

## 2.2   Why Generative AI Appeals to Users

From calculators, interfaces like spreadsheets, and modeling systems, to query languages, digital assistants, bots, and search, computers answer our questions in so many ways. Now we are enamored with the ease of solving problems with generative AI. What are elements that might contribute to generative AI being more accessible than other question-answering paradigms?

Generative AI has emerged as an easy way to form solutions. What are the elements that might contribute to it being more accessible than other question-answering paradigms?

1. Is it because proposed solutions are easier to follow than a taxonomic list?
2. Do people take naturally to the output because it shows the confidence of a proponent presenting a solution?
3. Is it because it answers with proposed solutions, not facts?
4. Is it the way it adds in connective plausible ideas that give smooth transitions? Are these "hallucinated" plausible ideas useful for reinforcing known facts? People also get away with adding such plausible but inaccurate ideas to connect facts and get away with them in promoting a narrative?
5. Is it the output language of formulating various kinds of prose structure that mimics the kinds we are used to people producing?
6. Do people take naturally to the input because people are accustomed to discourse?
7. Is the output valued because a story, right or wrong, is how people learn and remember?
8. Is it the way it presents ideas with a persuasive stance that draws us in as any promoting stance does?
9. Is it because the query can refine a previous result to make it even more appropriate?

Plausibly, all these factors help make generative AI attractive to users. But people using these tools also love to note limitations.

The average ChatGPT user recognizes that they can not find the source of the knowledge, but temper that with follow-up questions to guide the system to refine the answer. Generative AI can inform the user but the structure of the narrative comes off as formulaic and recognizable. They might still see these responses as a gift of something in the form they will need to produce, presented as an example they might even use. The scenario seems to speak to cognitive processes we are all familiar with from working with people. The results are presented in standard paragraphs and story structures we are used to. People recognize the results as well-produced first drafts. Generative AI might actually lead users to more fully consider results and use search more powerfully to get to solutions.

### 2.2.1   The Opportunities To Support People's Explorations and Expositions Are Vast

A quality idea-development exercise has many steps. Ideas come from a deep knowledge of many areas. The knowledgeable person, like repositories of online information, has representational and analytic knowledge of what they are working on. The "team" consists of people and AI-based knowledge resources that will have a range of solutions with purposeful idea development discussions.

Generative AI output responses being in specific coherent readable text softens incorrect responses with confident contextual style. With similar requests, today's generative AI makes lists that repeat or recognize style and structure. Still, the ability to follow up on a request is valuable. Discursive acts of every kind are supported by generative AI. The goal before us is to create a workflow that can verify, validate, and create a work product that results from a prompt.

Word-processing tools have revolutionized spell-checking, editing, layout typesetting, table of contents, and bibliography creation. Spreadsheets have organized simple analyses. Presentation creation tools have revolutionized organizing text, imagery, and video in support of a talk. Online search systems were designed to find information. generative AI was designed to produce plausible work products. The challenge is to integrate provenance and knowledge with its draft work products.

## 3   GAST - Generate And Search Test, A Knowledge Development Environment

It is my hope that we will make Knowledge Development Environment (KDE) tools that continue to improve people's ability and determination to solve physical, human, social, societal, governmental, and environmental problems.

The knowledge workers above have many goals, but all of them want to produce an answer or work product. In every case, they need to generate supported alternatives to complete their work. Generate And Search Test presents a scenario to do this.

We now have powerful tools for accessing and formulating ideas. In the past we thought bottom up; spend your life building the knowledge to get the right answer when you need it. Actually, for hard problems, no one instantly knows the answer. Different parts of creating a solution require different ways of working. We must try an idea and then test it. GAST puts forward a scenario that helps us use generative AI to letting us audition ideas that might help us get started. But people have been kicked out of classes and courts when they see ChatGPT as solutions, not just proposals. Any responsible result will only happen if we take steps to test and improve solutions. The below proposal creates a Knowledge Development Environment "war room" tool to build great solutions. GAST's goal is to give a problem-solving person better ways to consider solution value than any group of people has ever had.

The Generate And Search Test workflow combines knowledge tools and visualization with complementary strengths. Its goal is to help a person create and test hypothetical solutions in creating work products. GAST's goal is to give a problem-solving person better ways to consider solution value than any group of people has ever had.

It starts with a question that can be responded to with a narrative story, or as a simple answer with an information search, or a ChatGPT-like query. The user's goal should drive the process. One can be looking for a bibliographic reference, a fact, a person, a place, or a thing. Such might be best followed as a typical Boolean search query in a search box. Alternatively, the process can be an exploration that can be described with goals and a desired output format. The goal might be an explanation or to write a legal brief, a letter, or a research paper. To match the goal of the work product, the generative AI scenario encourages explanations about not just the question but the style of answer requested. Both search and generative AI approaches tend to get people to include follow-up queries.

The history of search has tended to be a list of stored searches in tabs or a series of links to be traversed. It can be made easier to keep track of the relationship between work products. To consider alternatives, the Generate And Search Test accumulates the query versions a user makes. These alternatives can be in collapsing columns that can be positioned next to each other for comparison. They can also have annotations or links to other things. A goal for a work product might give better ways to organize and collect these for comparison.

The result of the first request might look like the cognitive graphs that cognitive scientists have long promoted [13]. Alternatively, they might be presented as a graph in a mind map or outliner with links that let a person select parts of the generated texts for fact-checking search queries. Whether the results of these are shown as followable links that can pop up or are accessible in tabs or sidebars, new changes to the query or result should be visible without obscuring the solution. This becomes an expanding idea-deepening process.

The system moves beyond the exploration being the clutter of tabs in a browser we have been living with. GAST marks queries and results so they can be expandable to show what was followed and what was cast aside. GAST might:

1. identify or color the material to help the user follow steps to assure themselves they have understood how they did it.
2. be designed to integrate the many ways people explore information and generate solutions in text, imagery, and video.
3. show tested and untested ideas, edited and unedited material.
4. encourage users to improve idea support, test facts, try different scenarios, and consider solution value.

5. let users edit and add to audio, images, and videos they want to include.
6. produce evaluation metrics to help a user know how well they have done at assimilating the factual and artistic content.
7. have metrics for how much the content is owned by the user, generated by AI, or factually based.
8. be designed to be able to support users' needs in the process of sending vetted legal, personal, and professional work products.
9. Aid in exploring test plans, designs, and purchases.

GAST's approach creates an interface that helps people generate, test, and classify aspects of plausible work products as they are auditioned for the final chosen work product. A work product is a finished-looking narrative presentation backed up by a history of the knowledge generated-and-test journey. Exposing the journey creates and tests the solution in steps.

# 4 Broader uses of LLM AI

The last 20 years have shown the value of building systems that allow users to use keywords to access comprehensive knowledge sources. The value of these comprehensive data sources has changed the way we do everything, from play to work, to education, and to procurement.

Search has taken on many goals, creating its own websites, promoting goods and services, and helping people with directions and strategies. We come back to the point that a singular search box without user direction is probably not the final word on a knowledge interface. Search's many different functions should not seem overwhelming but might also benefit from differentiated search goals, as we see when we search in the "Google Scholar" search mode.

Decisions are always made in contexts. While complex interfaces can be overwhelming, part of the user experience for generative AI might become deploying reasonable search goals in context. The proposed Generate And Search Test tool above brings generative and search-based AI goals into an integrated solution-building scenario.

The work products we create with our knowledge are a different problem. The value of a story to learn and remember is well known. Generative AI also incorporates the context of recent requests in responses. To make a working product, a person needs to do things with facts. A story has to be believable and put into one's own words. Generate And Search Test is a tool to present a user-driven way that a knowledge worker can use AI. It posits that an integrated Knowledge Development Environment can manage to give an individual a broader ability than ever before to imagine and vet many solutions. It helps people produce results based on stories and facts.

The acquisition of facts has become fraught. Fake facts are found not just in generative AI's hallucinations, but also in things people make up or repeat to support or create some strategic social or political movement.

There used to be the idea that social media would help break down social barriers. It now seems that social media solving social problems was an early mirage.

Fantasies are stories rooted enough in reality that people imagine and often hope they will become reality. More than hallucinations, generative AI creates fantasies. At least parts of fantasies are, or can become, reality. Without checking out what in a fantasy is true or achievable, people can echo each other's fantasies in a spiral of confabulation. But communication and knowledge are important.

The current opportunities for using AI to teach and provoke people to create more expansive informed products are possibly the most important technical achievements of our time. It is up to us to make scenarios that improve the solutions people create.

An important step towards all of this is assured truth and provenance, two features that search has honored and continue to be critical for all real solutions. Generate And Search Test describes features for a workflow that will help people use confabulated solutions, not as fact, but as a way to creatively move from fantasies to scholarship-supported opportunities and solutions.

We look forward to a world in which peoples' searches, work, and communication are well-informed, incisive, and factual. I hope that Knowledge Development Environments (KDE) such as Generate and Search Test (GAST)can help us learn to make solutions to solve physical, human, social, societal, governmental, and environmental problems.

## ACKNOWLEDGMENTS

This paper was made possible with the generosity of Google. I especially appreciate the exciting talks and encouragement of Scott Penberthy, the editing help of Ellen Shay, Yunzi Wu, and Elizabeth Flynn, and comments of others.